# Lie Flow: Video Dynamic Fields Modeling and Predicting with Lie Algebra as Geometric Physics Principle

Weidong Qiao, Wangmeng Zuo, Hui Li*
Harbin Institute of Technology
wdqiao@stu.hit.edu.cn, cswmzuo@gmail.com, lihui@hit.edu.cn

## Abstract

*Modeling 4D scenes requires capturing both spatial structure and temporal motion, which is challenging due to the need for physically consistent representations of complex rigid and non-rigid motions. Existing approaches mainly rely on translational displacements, which struggle to represent rotations, articulated transformations, often leading to spatial inconsistency and physically implausible motion. LieFlow, a dynamic radiance representation framework that explicitly models motion within the SE(3) Lie group, enabling coherent learning of translation and rotation in a unified geometric space. The SE(3) transformation field enforces physically inspired constraints to maintain motion continuity and geometric consistency. The evaluation includes a synthetic dataset with rigid-body trajectories and two real-world datasets capturing complex motion under natural lighting and occlusions. Across all datasets, LieFlow consistently improves view-synthesis fidelity, temporal coherence, and physical realism over NeRF-based baselines. These results confirm that SE(3)-based motion modeling offers a robust and physically grounded framework for representing dynamic 4D scenes.*

## 1. Introduction

Motion is ubiquitous in the physical world. To compare the images and 3D scans describing the static structure of a scene, video provides a richer representation by capturing the trajectories, transformations, and interactions that characterize dynamic environments. Learning such motion dynamics from video has a wide real-world applications, including autonomous driving, human behavior analysis, and augmented/virtual reality. Consequently, modeling and predicting object motion from video has become a core problem in computer vision. Despite recent progress, building models that can faithfully capture both spatial and temporal dynamics of real-world scenes remains a formidable challenge, particularly when objects undergo complex, non-linear, or rigid-body transformations. This motivates the need for structured and physically grounded representations of motion, anchored in both geometry and time.

Recent advances in dynamic scene understanding have mainly focused on view synthesis and temporal extrapolation from monocular or multi-view videos. Existing approaches can be categorized into three classes. Time-parameterized methods enhance neural volumetric fields by incorporating time as a conditional input. However, they frequently conflate spatial and temporal variations, making it challenging to separate motion from static geometry. This entanglement limits their generalization capability, particularly in long-term predictions. Deformation-based methods learn per-point displacements, which restricts their ability to capture realistic motion patterns such as rotations, articulated transformations, or global rigid-body movements, and may result in spatially inconsistent deformations. Velocity or flow-based methods model motion through continuous trajectories or scene flows, which improve temporal coherence but often lack holistic structural constraints, leading to accumulated drift over time. Moreover, without structural priors, dense deformation fields can become spatially inconsistent or physically implausible.

In contrast to these paradigms, we propose a different framework that directly models dynamic fields as rigid-body transformations with the SE(3) Lie group because the Lie group can be used as an intrinsic geometrical physical principle. Instead of predicting dense flow fields or time-conditioned features, we represent motion using Lie algebra elements, which are mapped to SE(3) transformations via exponential mapping. This formulation naturally incorporates both translations and 3D rotations in a physically consistent and geometrically structured way. Our model can capture a broader and more complex range of motion patterns-encompassing not only local displacements but also global rotational dynamics and articulated rigid-body movements. This structured motion modeling enables better generalization across time and viewpoint and interpretability and controllability that are lost in deformation-based methods. The main contributions of this study are summarized as follows:

- We introduce an SE(3) transformation field grounded in Lie group theory, and provide a theoretical analysis to demonstrate its feasibility and effectiveness for modeling scene motions.

- We design a novel architecture, LieFlow, where the dynamic radiance field is represented by an enhanced HexPlane, and an SE(3) transformation field network captures frame-to-frame motion.
- We propose physics-inspired SE(3) constraints for dynamic modeling, including divergence-free regularization, momentum consistency, and group-structure preservation.
- We validate our method on both the synthetic dynamic object dataset and the real-world dynamic scene dataset, achieving superior performance in spatio-temporal novel view synthesis.

## 2. Related work

**Static 3D Representations:** Static 3D scene representations reconstruct geometry and appearance under fixed topology assumptions, forming the foundation for later dynamic models. Early approaches relied on explicit geometry, including voxel grids [1, 2, 3], polygonal meshes [4, 5, 6], and point-based structures [7, 8, 9, 10], which offered accurate geometry and real-time reconstruction but lacked photorealistic rendering and were memory-intensive, especially voxel-based ones [11, 12]. This limitation led to implicit representations, where continuous functions describe 3D geometry and appearance. Signed distance functions [13, 14], optimized via differentiable rendering [15, 16], jointly learn shape and appearance from 2D views, yet still struggle with high-fidelity view synthesis. NeRF [17] revolutionized photorealistic rendering by modeling volumetric density and radiance, with later variants improving quality [18, 19] and efficiency [20, 21]. Recently, 3D Gaussian Splatting [22] has become a compact explicit alternative offering real-time, high-fidelity rendering, supporting scalable reconstruction [23, 24, 25].

**Dynamic 3D Representations:** Real-world environments are inherently dynamic, as objects and scenes evolve continuously over time. Recent research has focused on extending 3D representations by explicitly incorporating the temporal dimension. Existing approaches to dynamic 3D representations can be broadly categorized into three main paradigms. Time-parameterized methods augment representations with time as an input, allowing implicit temporal modeling, widely used in dynamic NeRFs [26, 27] and Gaussian splatting [28, 29]. Deformation-based methods learn canonical spaces and deformation fields mapping to temporal states, as in D-NeRF [30], Nerfies [31], and HyperNeRF [32]; similar designs extend to Gaussian models such as 4D-GS [33], D3DGS [34], and MoSca [35]. Velocity or flow-based methods estimate motion fields or trajectories over time, enabling temporally consistent reconstructions in NSFF [36], DynNeRF [37], and NvFi [38]. Flow-guided Gaussian methods like SC-GS [39] and SplatFlow [40] track Gaussian primitives with velocity fields to capture fine inter-frame motion.

**Group Theory Deep Learning:** Group theory offers a principled framework for embedding geometric equivariance into neural networks by treating transformations as group operations. Early CNN-based works achieved discrete rotation and reflection equivariance via weight sharing [41, 42], later extended to continuous transformations using Lie groups, Lie algebras, and steerable filters [43, 44, 45, 46, 47, 48]. In 3D learning, group-based methods have been explored for pose tracking [49] and motion prediction [50]. Leveraging the SE(3) Lie group enables consistent motion modeling and invariant feature learning, enhancing temporal coherence and forming the basis of our method.

## 3. Method

### 3.1. Preliminaries

Motion in the physical world often arises from rigid-body dynamics, where objects preserve their geometry while undergoing translation and rotation. Accurately modeling such transformations is crucial for dynamic scene understanding. However, most existing dynamic radiance field methods rely on translation-only flow fields that independently displace each 3D point, failing to capture coherent rotational motion and resulting in spatially inconsistent deformations. To illustrate this limitation, consider a simple rigid-body motion around the z-axis. Let $\boldsymbol{p}_0 \in \mathbb{R}^3$ denote a 3D point on an object at time $t=0$. Suppose that at a future time $t$, the object undergoes a rotation $R_z(\theta)$ about the z-axis by an angle $\theta$, accompanied by a translation $\boldsymbol{t}_t \in \mathbb{R}^3$. The true rigid-body motion of the point is then

$$\boldsymbol{p}_t = R_z(\theta)\boldsymbol{p}_0 + \boldsymbol{t}_t \qquad (1)$$

If we attempt to approximate this motion using only a translation flow field $\boldsymbol{t}'_t$, the predicted motion trajectory becomes

$$\boldsymbol{p}'_t = \boldsymbol{p}_0 + \boldsymbol{t}'_t \qquad (2)$$

yielding a position error of

$$\begin{aligned}\epsilon_t &= \|\boldsymbol{p}_t - \boldsymbol{p}'_t\|_2^2 = \|R_z(\theta)\boldsymbol{p}_0 + \boldsymbol{t}_t - (\boldsymbol{p}_0 + \boldsymbol{t}'_t)\|_2^2 \\ &= \|(R_z(\theta) - I) \cdot \boldsymbol{p}_0 + (\boldsymbol{t}_t - \boldsymbol{t}'_t)\|_2^2\end{aligned} \qquad (3)$$

The first term reflects the rotation residual, which cannot be eliminated merely by optimizing the translation term $\boldsymbol{t}'_t$. As the object continues to move, this residual compounds and the modeling error increases.

To address these limitations, we adopt a more expressive and physically grounded representation – the Special Euclidean group denoted as SE(3), which unifies 3D rotation and translation in a continuous Lie group framework. Each rigid transformation is represented as

$$\text{SE}(3) = \left\{ g \mid g = \begin{bmatrix} R & t \\ 0 & 1 \end{bmatrix} \in \mathbb{R}^{4\times 4}, R \in SO(3), t \in \mathbb{R}^3 \right\} \quad (4)$$

where $R$ is a rotation matrix satisfying $R^T R = RR^T = I, |R| = 1$ and $t$ is the translation vector. The corresponding Lie algebra $\mathfrak{se}(3)$, parameterized by a 6D twist vector $\xi = \{\omega, v\}$, enables differentiable modeling through the exponential map

$$\exp(\hat{\xi}) = \sum_{n=0}^{\infty} \frac{\hat{\xi}^n}{n!} := \begin{bmatrix} R & t \\ 0 & 1 \end{bmatrix} \quad (5)$$

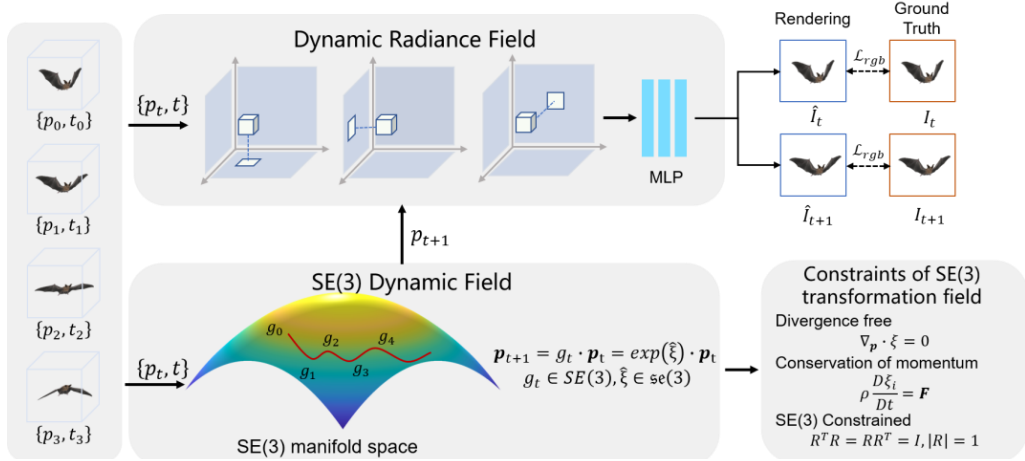

Figure 1: The overall method architecture.

The SE(3)-based motion model imposes strong geometric constraints (e.g., rotation matrix orthogonality), ensuring physical consistency and improving generalization in modeling complex real-world dynamics. The predicted rigid-body motion of a 3D point $p_0$ over time can thus be formulated as

$$\tilde{p}_t = g_t \cdot p_0 = exp(\hat{\xi}) \cdot p_0, g_t \in SE(3) \quad (6)$$

Therefore, the rigid body motion error of 3D objects predicted by the SE(3) Lie group is mainly determined by the integral accuracy of the Lie algebra.

### 3.2. Overview

We propose a general motion modeling framework that incorporates rigid-body transformations into dynamic scene understanding. Unlike previous methods tied to specific renderers, our approach learns motion in the Lie group SE(3), enabling broad compatibility with various neural representations. For demonstration, we adopt HexPlane [51] for its efficiency, though our SE(3)-based motion module can integrate with any backbone.

As shown in Figure 1, the overall architecture of our method framework consists of two key components: a dynamic radiance field network and an SE(3) transformation field. Given images with known poses, our model jointly learns time-varying radiance and motion. The dynamic radiance field encodes spatial features via HexPlane projections to predict density and color, while the SE(3) field predicts Lie algebra coefficients $\xi_t$ from the current spatial point $p_t$, timestamp $t$, and time interval $\Delta t$, which is applied to warp the point to the next time step $p_{t+1}$. To ensure physically plausible motion, we regularize the SE(3) field with divergence-free, momentum-conservation, and rotation-orthogonality constraints.

### 3.3. Optimization of the dynamic radiance field

We adopt a dynamic radiance field architecture based on a modified HexPlane representation. This design enables a compact, multi-scale, and geometry-aware embedding of spatiotemporal information, supporting fast convergence and generalization across diverse dynamic scenes.

The detailed network architecture of the dynamic radiance field is illustrated in Fig. 2. The network takes a 3D point $p_t \in \mathbb{R}^3$ and a normalized timestamp $t \in [0,1]$, and outputs the corresponding radiance color and volume density. The inputs are projected onto six learnable 2D feature planes along three spatial axes $(x,y), (x,z), (y,z)$ and three spatiotemporal axes $(z,t), (y,t), (x,t)$. Each projection is sampled from a fixed-resolution tensor (e.g., 128×128), which is sampled from learnable tensor fields using bilinear interpolation. This forms the space-time separable encoding of the point. Then, the extracted features from the spatial and temporal planes are multiplied (element-wise) and passed through two separate linear layers, one maps the density-related features to a scalar volume density $\sigma$, while the other maps the appearance-related features to a latent color embedding. A rendering module takes this embedding, along with the viewing direction, and predicts the final RGB color of the point. The final color is accumulated along camera rays using standard volume rendering.

During training, the network is supervised by the photometric reconstruction loss between the rendered image and the ground-truth observation. Specifically, for each camera ray r, the loss is defined as:

$$\mathcal{L}_{pho} = \left\| C_t(r) - C_t^{gt}(r) \right\| \quad (7)$$

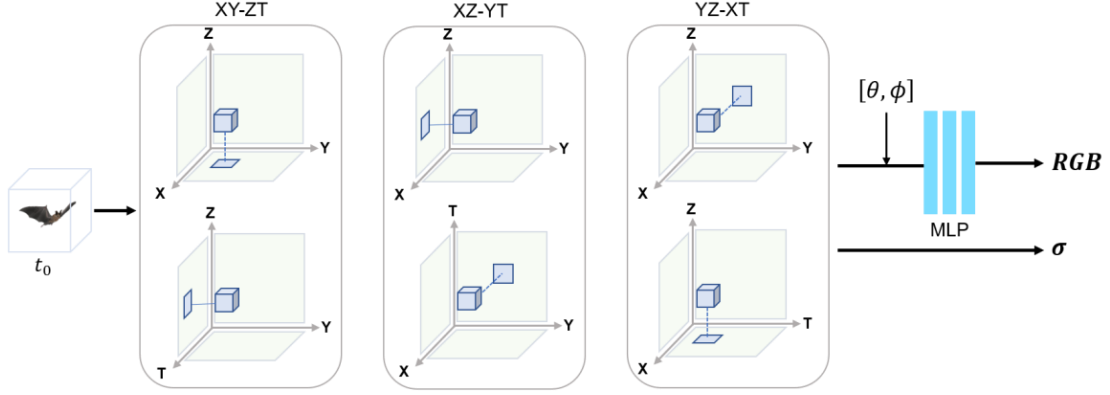

Figure 2: The overall dynamic radiance field architecture.

where $C_t(r)$ is the rendered color along the camera ray $r$ at time $t$, and $C_t^{gt}(r)$ denotes the corresponding ground-truth pixel value.

### 3.4. Optimization of SE(3) transformation field

The overall structure of the SE(3) transformation field network is illustrated in Fig. 3. This network is to learn a continuous, rigid transformation function that maps any point $p_t \in \mathbb{R}^3$ at time $t \in [0,1]$ to its corresponding canonical space representation via elements of the SE(3) Lie group. These inputs are first processed by a Position and Time Encoder that applies fixed sinusoidal positional encoding using sin and cos functions at multiple frequencies, enabling the network to capture fine-grained spatial and temporal variations. The encoded vector is then fed into two parallel MLP branches, one of which outputs a 3D angular velocity vector $\boldsymbol{\omega}_t = (\omega_x, \omega_y, \omega_z)^T$, representing rotation, another branch outputs a 3D translational velocity $\boldsymbol{v}_t = (v_x, v_y, v_z)^T$. Each MLP consists of 4 layers with 128 hidden units. These two outputs are concatenated into a 6D twist vector $\boldsymbol{\xi}_t = [\boldsymbol{\omega}_t, \boldsymbol{v}_t] \in \mathbb{R}^6$, which lies in the Lie algebra $\mathfrak{se}(3)$. Finally, the matrix exponential is applied to map $\boldsymbol{\xi}_t$ to the SE(3) Lie group $g_t = \exp(\hat{\boldsymbol{\xi}}_t) \in SE(3)$. The resulting rigid transformation $g_t$ is then applied to the point $p_t$ to obtain its canonical representation $\tilde{p}_t = g_t \cdot p_t$. The use of the SE(3) Lie group structure ensures the differentiability and smoothness of the transformation field. This transformation field is queried at every spatial location and time step during training and serves as the motion-warping function for rendering and supervision.

Rather than warping all points to a single canonical time (e.g., t = 0), which can result in long-range transformations and unstable optimization, we adopt a sparse reference frame strategy. Specifically, we select a subset of frames (e.g., every 4th frame) as reference frames, and define the radiance field in their coordinate systems, enabling localized canonical space across time as

$$(c, \sigma) = f(x, y, z, t_k), t_k \in \{t | t \bmod 4 = 0\} \quad (8)$$

where $f$ means the dynamic radiance field network and $t_k$ denotes the reference frames. All other query frames are transformed to the nearest reference frame via integration over the SE(3) transformation field as

$$\boldsymbol{\xi} = [\boldsymbol{\omega}, \boldsymbol{v}]^T = g(x, y, z, t_i),$$
$$t_i \in \{t | t \bmod 4 \neq 0\} \quad (9)$$

where $g$ represents the SE(3) transformation field network and $t_i$ denotes the query frames. This strategy reduces the length of transformation paths, preventing distortions caused by integrating over long temporal intervals, and ensures better temporal continuity.

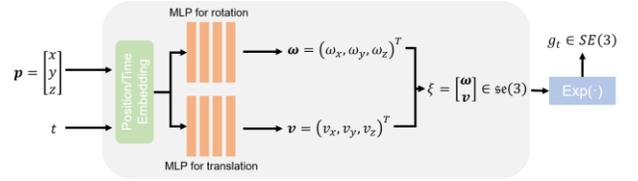

Figure 3: The overall SE(3) transformation field network architecture.

To render an image at a non-reference time $t_i$, a coordinate transformation strategy that maps all sampled points to the nearest reference frame for querying is employed. Given a 3D point $p_s$ sampled along a camera ray at a query time $t_i$, we first determine the nearest reference timestamp $\hat{t}_k$ such that

$$\hat{t}_k = \underset{t_k \in Ref}{\arg\min} |t_k - t_i| \quad (10)$$

Then the SE(3) transformation field network $g$ integrates over the temporal interval $[t_i, \hat{t}_k]$, producing a continuous tangent vector $\boldsymbol{\xi}(\tau) = g(p_s, \tau) \in \mathfrak{se}(3)$, which defines the instantaneous rigid motion at each timestep. The full transformation matrix is obtained by integrating this Lie algebra field

$$G(t_i \to \hat{t}_k) = exp\left(\int_{t_i}^{\hat{t}_k} \boldsymbol{\xi}(\tau) d\tau\right) \quad (11)$$

This transformation warps the sampled point from its

original location to the reference coordinate system
$$\widetilde{\boldsymbol{p}}_s = G(t_i \to \hat{t}_k) \cdot \boldsymbol{p}_s \tag{12}$$
The warped point $\widetilde{\boldsymbol{p}}_s$ is then queried in the canonical radiance field defined at time $\hat{t}_k$, producing a volume density $\sigma_s$ and a view-independent appearance embedding $e_s$. To maintain temporal consistency, the final color prediction is performed using the original viewing direction from time $t_i$
$$\boldsymbol{c}_s = f(e_s, \boldsymbol{v}_i) \tag{13}$$
where $\boldsymbol{v}_i$ is the ray direction. The rendered image at $t_i$ is synthesized using volumetric rendering along the ray. This process is illustrated in Fig. 4, where a point in the query frame is transformed to the reference coordinate system via the learned SE(3) transformation field and used for querying the canonical field before rendering. This approach enables temporally aligned supervision without requiring separate field definitions for every frame.

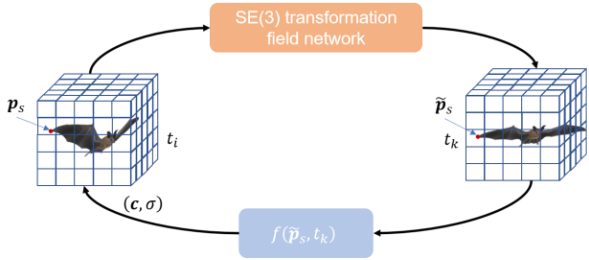

Figure 4: Rendering via transformation to canonical space.

A set of physically inspired and structure-aware loss terms on the transformation field network $g$ are imposed to ensure physically plausible and geometrically valid SE(3) transformations. These consist of divergence-free flow regularization, momentum conservation, and SE(3) structural constraints. To prevent spatial expansion or collapse in the predicted motion field, the SE(3) transformation field $\xi(\boldsymbol{p}, t)$ should be divergence-free
$$\nabla_p \cdot \xi = 0 \tag{14}$$
The corresponding loss is
$$\mathcal{L}_{divergence\_free} = \frac{1}{NM} \sum_{n=1}^{N} \sum_{m=1}^{M} \|\nabla_{p_n} \cdot \xi(\boldsymbol{p}_n, t_m)\| \tag{15}$$
Inspired by physical laws, the conservation of momentum using the material derivative of the transformation field
$$\frac{D\xi}{Dt} = \frac{\partial \xi}{\partial t} + \xi \cdot \nabla \xi = \boldsymbol{a} \tag{16}$$
where $\boldsymbol{a}$ is a learnable acceleration prior. The momentum loss is defined as
$$\mathcal{L}_{momentum} = \frac{1}{NM} \sum_{n=1}^{N} \sum_{m=1}^{M} \left\| \frac{\partial \xi(\boldsymbol{p}_n, t_m)}{t_m} + \xi(\boldsymbol{p}_n, t_m) \right. \\ \left. \cdot \nabla_{p_n} \xi(\boldsymbol{p}_n, t_m) - \boldsymbol{a} \right\| \tag{17}$$

We also explicitly regularize the predicted SE(3) matrices to remain close to the Lie group. The rotation matrices $R_i$ are encouraged to be orthogonal
$$\mathcal{L}_{ortho} = \frac{1}{N} \sum_{i=1}^{N} \|R_i R_i^T - I\| \tag{18}$$
and the translation vectors $\boldsymbol{v}$ are regularized for temporal smoothness
$$\mathcal{L}_{trans} = \frac{1}{N} \sum_{i=1}^{N} \|\boldsymbol{v}\| \tag{19}$$
The full SE(3) transformation regularization loss is
$$\mathcal{L}_{SE3} = \mathcal{L}_{ortho} + \mathcal{L}_{trans} \tag{20}$$
Finally, the SE(3) transformation field network $g$ is trained under the combined supervision of physical and geometric constraints as well as query frame photometric loss
$$g \leftarrow (\mathcal{L}_{divergence\_free} + \mathcal{L}_{momentum} + \mathcal{L}_{pho\_query} + \mathcal{L}_{SE3}) \tag{21}$$
Meanwhile, the radiance field network $f$ is optimized with reference frames and query frames photometric losses
$$f \leftarrow \mathcal{L}_{pho\_ref} + \mathcal{L}_{pho\_query} \tag{22}$$

## 4. Experiments

### 4.1. Experimental setup

To evaluate our method's effectiveness in modeling rigid-body transformations of dynamic objects, we conduct experiments on two challenging dynamic scene datasets. These datasets assess novel view synthesis across spatial and temporal domains for both interpolation and extrapolation. The synthetic dataset provides clean geometry with physical motion, while the real-world dataset captures complex dynamics under natural conditions. Training is performed on two NVIDIA A6000 GPUs, and further details are presented in the following subsections.

### 4.2. Results on the synthetic dynamic object dataset

The synthetic dynamic object dataset [38] contains six animated 3D object scenes exhibiting diverse rigid and non-rigid motions. Each scene is rendered from 15 viewpoints over 1 second at 60 fps. We use 12 viewpoints for training and the rest for interpolation and extrapolation evaluation. Our method is compared with D-NeRF [30], TiNeuVox [52], NvFi [38], and SC-GS [39], using PSNR, SSIM, and LPIPS as metrics. As shown in Table 1, our method achieves the highest average performance across all tasks, confirming its strong capability in modeling complex 3D motion for high-quality novel view synthesis.

Table 1 compares all methods in terms of view synthesis quality on the six animated 3D object scenes, for both interpolation and extrapolation tasks. Our method achieves the highest average performance across all tasks, demonstrating its strong ability to capture and generalize complex rigid-body motions for high-quality novel view synthesis in both spatial and temporal dimensions.

Table 1 Comparison results of methods in view synthesis quality on the synthetic dynamic object dataset.

| Methods | Interpolation | | | Extrapolation | | |
|---|---|---|---|---|---|---|
| | PSNR | SSIM | LPIPS | PSNR | SSIM | LPIPS |
| D-NeRF | 14.158 | 0.697 | 0.352 | 14.660 | 0.737 | 0.312 |
| TiNeuVox | 27.988 | 0.960 | 0.063 | 19.612 | 0.940 | 0.073 |
| NvFi | 29.027 | 0.970 | 0.039 | 27.594 | **0.972** | **0.036** |
| SC-GS | 9.046 | 0.839 | 0.141 | 8.977 | 0.846 | 0.135 |
| Ours | **30.802** | **0.975** | **0.036** | **28.141** | 0.972 | 0.037 |

Figure 5 presents qualitative comparisons of the Fan and Whale sequences. Our method produces visually clean results, with backgrounds free from noticeable artifacts or residual noise. The dynamic objects are rendered with high fidelity – the fan blades exhibit precise rotational motion, while the whale's tail demonstrates smooth and accurate up-down movement. These results further confirm that our explicit SE(3) transformation modeling can faithfully capture complex dynamic motions over time.

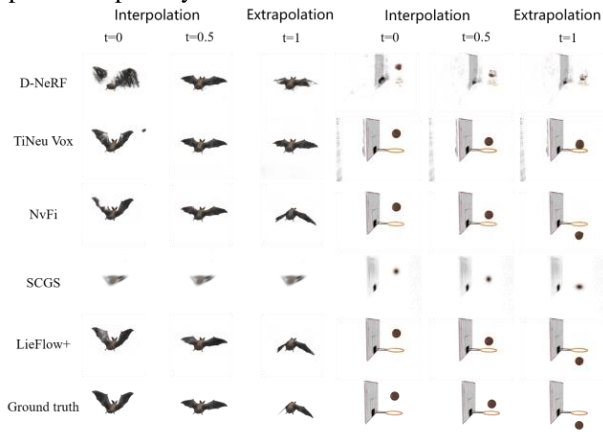

Figure 5: Comparisons of multiple methods on the Bat and FallingBall sequences.

### 4.3. Results on the NVIDIA dynamic scene dataset

We further evaluate our method on the NVIDIA Dynamic Scene Dataset [53], which contains real-world multi-view videos of human actions captured by 12 synchronized cameras. To simulate monocular camera motion, one frame per time step is selected from different viewpoints. Each scene includes a static background and dynamic foreground with provided ground-truth depth, flow, and masks.

To handle the interference between static and dynamic regions, we adopt a foreground–background disentanglement strategy using object masks. The background field is modeled as time-invariant, while the foreground field captures SE(3) motion. This separation improves both background reconstruction and dynamic motion learning.

Table 2 compares our method with NeRF+Time, DynNeRF [37], SC-GS [39], and MoSca [35] on four representative sequences. Our method achieves superior average PSNR and LPIPS performance, demonstrating its effectiveness in modeling real-world dynamic scenes with complex motion.

As shown in Table 2, our method achieves consistently superior performance across all four sequences. LieFlow attains the highest average score (25.73/0.051), showing both sharper reconstruction quality and better perceptual fidelity across rigid (Balloon1, Balloon2, Playground) and non-rigid (Umbrella) motions. These results validate the effectiveness of our SE(3) transformation field in modeling real-world dynamics.

Table 2 Comparison results of methods in view synthesis on the NVIDIA dataset.

| Methods | balloon1 PSNR/ LPIPS | balloon2 PSNR/ LPIPS | playground PSNR/ LPIPS | umbrella PSNR/ LPIPS | average PSNR/ LPIPS |
|---|---|---|---|---|---|
| NeRF+time | 15.33/ 0.304 | 17.67/ 0.236 | 13.80/ 0.444 | 15.17/ 0.752 | 15.49/ 0.434 |
| DynNeRF | 21.47/ 0.125 | 25.97/ 0.059 | 23.65/ 0.093 | 23.15/ 0.146 | 23.56/ 0.106 |
| SC-GS | 20.17/ 0.179 | 21.07/ 0.149 | 20.71/ 0.115 | 21.84/ 0.160 | 20.95/ 0.151 |
| MoSca | 23.26/ 0.092 | 28.90/ 0.042 | 23.05/ 0.060 | 24.41/ 0.092 | 24.90/ 0.072 |
| Ours | **23.59/ 0.062** | **28.93/ 0.026** | **24.84/ 0.051** | **25.57/ 0.066** | **25.73/ 0.051** |

The visual comparisons of representative scenes from the NVIDIA Dynamic Scene dataset are presented in Figure 6. As shown in Figure 6, our method produces sharper and more realistic reconstructions of both static background and dynamic foreground objects. In particular, the fine details of object boundaries and the consistency of motion over time are better preserved in our results. These visual comparisons demonstrate the ability of our SE(3) transformation field to capture complex motions in real-world dynamics scenes.

### 4.4. Qualitative evaluation

We further perform qualitative comparisons on selected sequences from the DAVIS dataset (swing, train, horsejump-high, horsejump-low), using STGS [54] and RoDynRF [55] as representative baselines. As shown in Figure 7, our method is capable of generating high-quality views that accurately recover the motion and appearance of dynamic objects. The STGS fails to produce valid results on the DAVIS dataset because COLMAP cannot generate reliable sparse point clouds or camera parameters under its monocular setting. In contrast, our method directly leverages the visual cues already present in the video frames and reconstructs the scene without relying on precomputed geometry or multi-view calibration. The SE(3) transformation field framework effectively models both object motion and viewpoint variation for view

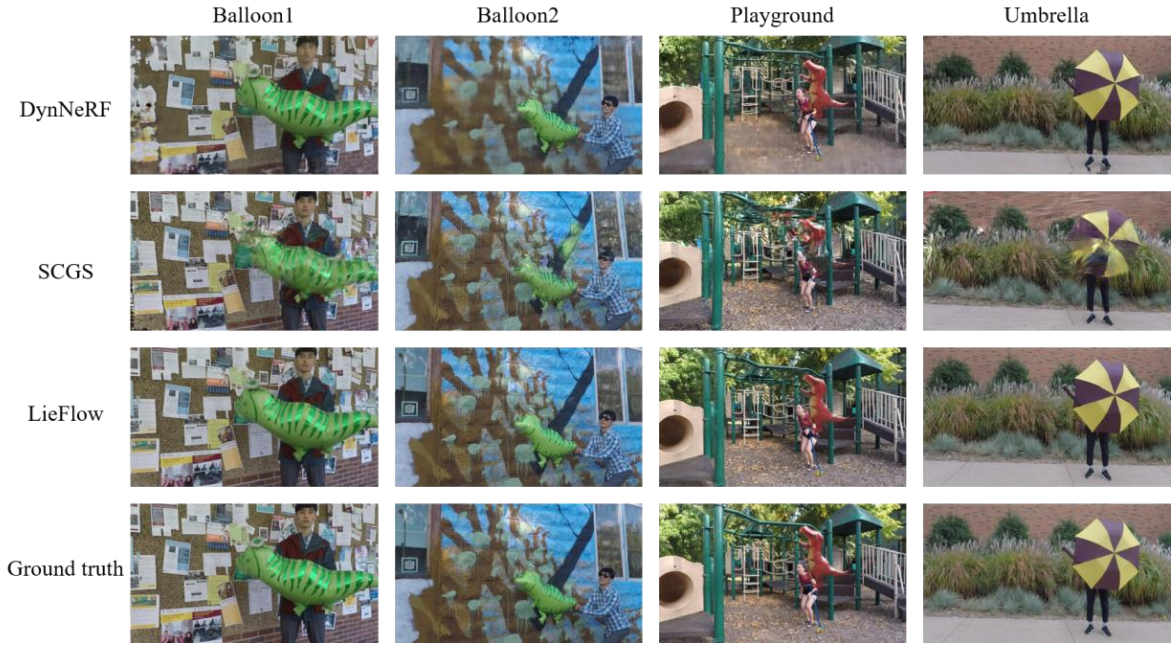

Figure 6: Visual comparisons of multiple methods on the NVIDIA dynamic scene dataset.

synthesis from purely monocular inputs.

### 4.5. Ablation study

To assess the necessity of full SE(3) modeling, we conduct an ablation study comparing translation-only, rotation-only, and full variants of our framework on the synthetic dynamic object dataset. Each variant is trained under the same protocol.

As shown in Table 3, the full SE(3) transformation field achieves comparable interpolation accuracy but significantly outperforms the single-component variants in extrapolation. This indicates that rigid motion cannot be effectively represented by translation or rotation alone. Joint modeling of both components provides more accurate and stable motion representations, confirming the importance of SE(3) formulation for dynamic scene synthesis.

Table 3 Quantitative results of ablation studies on the synthetic dynamic object dataset.

| Transformation type | Interpolation | | | Extrapolation | | |
|---|---|---|---|---|---|---|
| | PSNR | SSIM | LPIPS | PSNR | SSIM | LPIPS |
| Translation field | 30.438 | 0.976 | 0.034 | 26.457 | 0.968 | 0.038 |
| Rotation field | 30.144 | 0.974 | 0.036 | 24.111 | 0.958 | 0.044 |
| SE(3) field | **30.802** | 0.975 | 0.036 | **28.141** | **0.972** | 0.037 |

Figure 7 presents qualitative comparisons of synthesized frames under different transformation field variants. We observe that results obtained using only the translation field often lack structural completeness, since rotational consistency is missing and object parts become distorted. Similarly, using only the rotation field fails to capture the global displacement of objects, as global translations are not properly captured. In contrast, SE(3) field effectively combines both translation and rotation, producing results in which the dynamic objects move into their correct spatial locations while preserving structural integrity. These observations highlight the importance of modeling rigid motion holistically with SE(3) transformations rather than relying on isolated components.

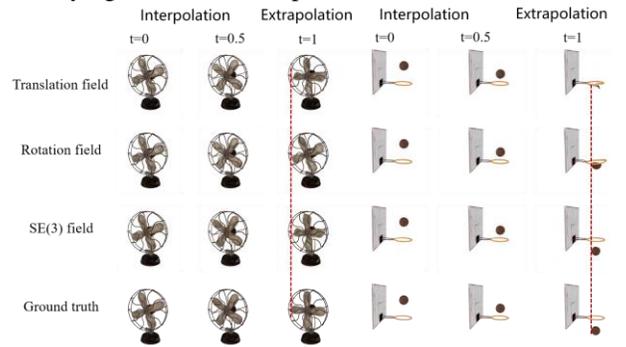

Figure 7. Comparisons of multiple transformation types on the synthetic dynamic object dataset.

### 5. Conclusion

This study presents a LieFlow, a novel framework for dynamic 3D scene representation that introduces an SE(3)-based transformation field grounded in Lie group theory. By combining an improved HexPlane with an SE(3) motion modeling network, and incorporating physically inspired constraints such as divergence-free regularization, momentum conservation, and intrinsic SE(3) consistency, our method effectively captures both rotational and translational dynamics. Experimental evaluations on synthetic and real-world benchmarks demonstrate that LieFLow achieves consistent improvements in spatio-temporal novel view synthesis, confirming the effectiveness of SE(3)-based motion modeling.

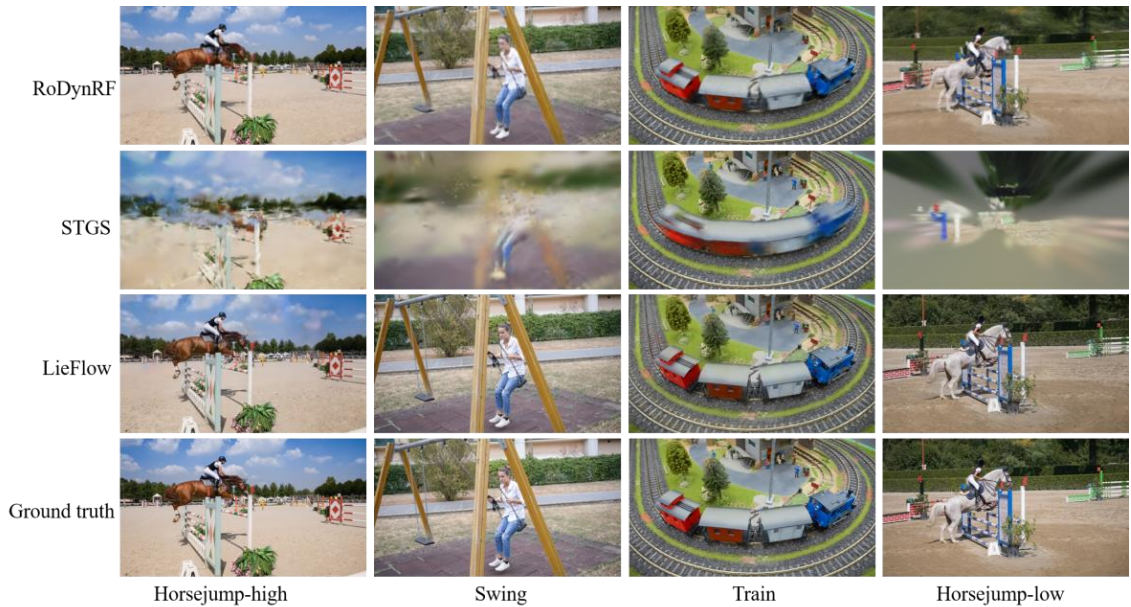

Figure 7: Qualitative comparisons on the DAVIS dataset.

In future research, we plan to extend SE(3) transformation fields as a generalizable plug-in module for a broader range of dynamic 3D representation methods. This will further validate the universality and flexibility of our approach, and potentially advance the robustness and scalability of motion modeling in complex dynamic scenes. Moreover, we aim to explore non-rigid motion modeling by integrating an appropriate Lie group or deformation-based representations to further enhance the capability of capturing complex dynamics in real-world scenes.

## Acknowledgments

The authors would like to acknowledge the National Natural Science Foundation of China (Grant Nos. 52441803).

## References

[1] Wu, Z., Song, S., Khosla, A., Yu, F., Zhang, L., Tang, X., & Xiao, J. (2015). 3d shapenets: A deep representation for volumetric shapes. In *Proceedings of the IEEE Conference on Computer Vision and Pattern Recognition* (pp. 1912-1920).

[2] Zhou, Y., & Tuzel, O. (2018). Voxelnet: End-to-end learning for point cloud based 3d object detection. In *Proceedings of the IEEE Conference on Computer Vision and Pattern Recognition* (pp. 4490-4499).

[3] Peng, K., Islam, R., Quarles, J., & Desai, K. (2022). Tmvnet: Using transformers for multi-view voxel-based 3d reconstruction. In *Proceedings of the IEEE/CVF Conference on Computer Vision and Pattern Recognition* (pp. 222-230).

[4] Kazhdan, M., Bolitho, M., & Hoppe, H. (2006). Poisson surface reconstruction. In *Proceedings of the Fourth Eurographics Symposium on Geometry Processing* (Vol. 7, No. 4).

[5] Wang, N., Zhang, Y., Li, Z., Fu, Y., Liu, W., & Jiang, Y. G. (2018). Pixel2mesh: Generating 3d mesh models from single rgb images. In *Proceedings of the European Conference on Computer Vision (ECCV)* (pp. 52-67).

[6] Vakalopoulou, M., Chassagnon, G., Bus, N., Marini, R., Zacharaki, E. I., Revel, M. P., & Paragios, N. (2018). Atlasnet: Multi-atlas non-linear deep networks for medical image segmentation. In *International Conference on Medical Image Computing and Computer-Assisted Intervention* (pp. 658-666).

[7] Qi, C. R., Su, H., Mo, K., & Guibas, L. J. (2017). Pointnet: Deep learning on point sets for 3d classification and segmentation. In *Proceedings of the IEEE Conference on Computer Vision and Pattern Recognition* (pp. 652-660).

[8] Qi, C. R., Yi, L., Su, H., & Guibas, L. J. (2017). Pointnet++: Deep hierarchical feature learning on point sets in a metric space. *Advances in Neural Information Processing Systems*, *30*.

[9] Wang, Y., Sun, Y., Liu, Z., Sarma, S. E., Bronstein, M. M., & Solomon, J. M. (2019). Dynamic graph cnn for learning on point clouds. *ACM Transactions on Graphics*, *38*(5), 1-12.

[10] Zhao, H., Jiang, L., Jia, J., Torr, P. H., & Koltun, V. (2021). Point transformer. In *Proceedings of the IEEE/CVF International Conference on Computer Vision* (pp. 16259-16268).

[11] Nießner, M., Zollhöfer, M., Izadi, S., & Stamminger, M. (2013). Real-time 3D reconstruction at scale using voxel hashing. *ACM Transactions on Graphics*, *32*(6), 1-11.

[12] Choy, C. B., Xu, D., Gwak, J., Chen, K., & Savarese, S. (2016). 3d-r2n2: A unified approach for single and multi-view 3d object reconstruction. In *European Conference on Computer Vision* (pp. 628-644).

[13] Park, J. J., Florence, P., Straub, J., Newcombe, R., & Lovegrove, S. (2019). Deepsdf: Learning continuous signed distance functions for shape representation. In *Proceedings of the IEEE/CVF Conference on Computer Vision and Pattern Recognition* (pp. 165-174).

[14] Yariv, L., Kasten, Y., Moran, D., Galun, M., Atzmon, M., Ronen, B., & Lipman, Y. (2020). Multiview neural surface reconstruction by disentangling geometry and appearance. *Advances in Neural Information Processing Systems*, *33*, 2492-2502.

[15] Loper, M. M., & Black, M. J. (2014). OpenDR: An approximate differentiable renderer. In *European Conference on Computer Vision* (pp. 154-169).

[16] Liu, S., Saito, S., Chen, W., & Li, H. (2019). Learning to infer implicit surfaces without 3d supervision. *Advances in Neural Information Processing Systems*, *32*.

[17] Mildenhall, B., Srinivasan, P. P., Tancik, M., Barron, J. T., Ramamoorthi, R., & Ng, R. (2021). Nerf: Representing scenes as neural radiance fields for view synthesis. *Communications of the ACM*, *65*(1), 99-106.

[18] Barron, J. T., Mildenhall, B., Tancik, M., Hedman, P., Martin-Brualla, R., & Srinivasan, P. P. (2021). Mip-nerf: A multiscale representation for anti-aliasing neural radiance fields. In *Proceedings of the IEEE/CVF International Conference on Computer Vision* (pp. 5855-5864).

[19] Zhang, K., Riegler, G., Snavely, N., & Koltun, V. (2020). Nerf++: Analyzing and improving neural radiance fields. *arXiv:2010.07492*.

[20] Müller, T., Evans, A., Schied, C., & Keller, A. (2022). Instant neural graphics primitives with a multiresolution hash encoding. *ACM Transactions on Graphics*, *41*(4), 1-15.

[21] Chen, A., Xu, Z., Geiger, A., Yu, J., & Su, H. (2022). Tensorf: Tensorial radiance fields. In *European Conference on Computer Vision* (pp. 333-350).

[22] Kerbl, B., Kopanas, G., Leimkühler, T., & Drettakis, G. (2023). 3D Gaussian splatting for real-time radiance field rendering. *ACM Trans. Graph.*, *42*(4), 139-1.

[23] Chen, Y., Xu, H., Zheng, C., Zhuang, B., Pollefeys, M., Geiger, A., Cham, T., & Cai, J. (2024). Mvsplat: Efficient 3d gaussian splatting from sparse multi-view images. In *European Conference on Computer Vision* (pp. 370-386).

[24] Lin, J., Li, Z., Tang, X., Liu, J., Liu, S., Liu, J., Lu, Y., Wu, X., Xu, S., Yan, Y., & Yang, W. (2024). Vastgaussian: Vast 3d gaussians for large scene reconstruction. In *Proceedings of the IEEE/CVF Conference on Computer Vision and Pattern Recognition* (pp. 5166-5175).

[25] Peng, Z., Shao, T., Liu, Y., Zhou, J., Yang, Y., Wang, J., & Zhou, K. (2024). Rtg-slam: Real-time 3d reconstruction at scale using gaussian splatting. In *ACM SIGGRAPH 2024 Conference Papers* (pp. 1-11).

[26] Attal, B., Laidlaw, E., Gokaslan, A., Kim, C., Richardt, C., Tompkin, J., & O'Toole, M. (2021). Törf: Time-of-flight radiance fields for dynamic scene view synthesis. *Advances in Neural Information Processing Systems*, *34*, 26289-26301.

[27] Li, T., Slavcheva, M., Zollhoefer, M., Green, S., Lassner, C., Kim, C., Schmidt, T., Lovegrove, S., Goesele, M., Newcombe, R., & Lv, Z. (2022). Neural 3d video synthesis from multi-view video. In *Proceedings of the IEEE/CVF conference on computer vision and pattern recognition* (pp. 5521-5531).

[28] Katsumata, K., Vo, D. M., & Nakayama, H. (2024). A compact dynamic 3d gaussian representation for real-time dynamic view synthesis. In *European Conference on Computer Vision* (pp. 394-412).

[29] Li, Z., Chen, Z., Li, Z., & Xu, Y. (2024). Spacetime gaussian feature splatting for real-time dynamic view synthesis. In *Proceedings of the IEEE/CVF Conference on Computer Vision and Pattern Recognition* (pp. 8508-8520).

[30] Pumarola, A., Corona, E., Pons-Moll, G., & Moreno-Noguer, F. (2021). D-nerf: Neural radiance fields for dynamic scenes. In *Proceedings of the IEEE/CVF Conference on Computer Vision and Pattern Recognition* (pp. 10318-10327).

[31] Park, K., Sinha, U., Barron, J. T., Bouaziz, S., Goldman, D. B., Seitz, S. M., & Martin-Brualla, R. (2021). Nerfies: Deformable neural radiance fields. In *Proceedings of the IEEE/CVF International Conference on Computer Vision* (pp. 5865-5874).

[32] Park, K., Sinha, U., Hedman, P., Barron, J. T., Bouaziz, S., Goldman, D. B., Brualla, R. M., & Seitz, S. M. (2021). Hypernerf: A higher-dimensional representation for topologically varying neural radiance fields. *arXiv:2106.13228*.

[33] Wu, G., Yi, T., Fang, J., Xie, L., Zhang, X., Wei, W., Liu, W., & Wang, X. (2024). 4d gaussian splatting for real-time dynamic scene rendering. In *Proceedings of the IEEE/CVF Conference on Computer Vision and Pattern Recognition* (pp. 20310-20320).

[34] Yang, Z., Gao, X., Zhou, W., Jiao, S., Zhang, Y., & Jin, X. (2024). Deformable 3d gaussians for high-fidelity monocular dynamic scene reconstruction. In *Proceedings of the IEEE/CVF Conference on Computer Vision and Pattern Recognition* (pp. 20331-20341).

[35] Lei, J., Weng, Y., Harley, A. W., Guibas, L., & Daniilidis, K. (2025). Mosca: Dynamic gaussian fusion from casual videos via 4d motion scaffolds. In *Proceedings of the Computer Vision and Pattern Recognition Conference* (pp. 6165-6177).

[36] Li, Z., Niklaus, S., Snavely, N., & Wang, O. (2021). Neural scene flow fields for space-time view synthesis of dynamic scenes. In *Proceedings of the IEEE/CVF Conference on Computer Vision and Pattern Recognition* (pp. 6498-6508).

[37] Gao, C., Saraf, A., Kopf, J., & Huang, J. B. (2021). Dynamic view synthesis from dynamic monocular video. In *Proceedings of the IEEE/CVF International Conference on Computer Vision* (pp. 5712-5721).

[38] Li, J., Song, Z., & Yang, B. (2023). Nvfi: Neural velocity fields for 3d physics learning from dynamic videos. *Advances in Neural Information Processing Systems*, *36*, 34723-34751.

[39] Huang, Y. H., Sun, Y. T., Yang, Z., Lyu, X., Cao, Y. P., & Qi, X. (2024). Sc-gs: Sparse-controlled gaussian splatting for editable dynamic scenes. In *Proceedings of the IEEE/CVF Conference on Computer Vision and Pattern Recognition* (pp. 4220-4230).

[40] Wang, B., Zhang, Y., Li, J., Yu, Y., Sun, Z., Liu, L., & Hu, D. (2024). SplatFlow: Learning multi-frame optical flow via splatting. *International Journal of Computer Vision*, *132*(8), 3023-3045.

[41] Cohen, T., & Welling, M. (2016). Group equivariant convolutional networks. In *International Conference on Machine Learning* (pp. 2990-2999).

[42] Ravanbakhsh, S., Schneider, J., & Poczos, B. (2017). Equivariance through parameter-sharing. In *International Conference on Machine Learning* (pp. 2892-2901).

[43] Cohen, T. S., Geiger, M., & Weiler, M. (2019). A general theory of equivariant cnns on homogeneous spaces. *Advances in Neural Information Processing Systems*, *32*.

[44] Finzi, M., Stanton, S., Izmailov, P., & Wilson, A. G. (2020). Generalizing convolutional neural networks for equivariance to lie groups on arbitrary continuous data. In *International Conference on Machine Learning* (pp. 3165-3176).

[45] Bekkers, E. J. (2019). B-spline cnns on lie groups. arXiv:1909.12057.

[46] Sosnovik, I., Szmaja, M., & Smeulders, A. (2019). Scale-equivariant steerable networks. arXiv:1910.11093.

[47] Dehmamy, N., Walters, R., Liu, Y., Wang, D., & Yu, R. (2021). Automatic symmetry discovery with lie algebra convolutional network. *Advances in Neural Information Processing Systems, 34*, 2503-2515.

[48] Qiao, W., Xu, Y., & Li, H. (2025). Lie group convolution neural networks with scale-rotation equivariance. *Neural Networks*, *183*, 106980.

[49] Lin, J., Li, H., Chen, K., Lu, J., & Jia, K. (2021). Sparse steerable convolutions: An efficient learning of se (3)-equivariant features for estimation and tracking of object poses in 3d space. *Advances in Neural Information Processing Systems*, *34*, 16779-16790.

[50] Zaier, M., Wannous, H., & Drira, H. (2025). Geometry-Aware Deep Learning for 3D Skeleton-Based Motion Prediction. In *2025 IEEE/CVF Winter Conference on Applications of Computer Vision (WACV)* (pp. 4831-4840).

[51] Cao, A., & Johnson, J. (2023). Hexplane: A fast representation for dynamic scenes. In *Proceedings of the IEEE/CVF Conference on Computer Vision and Pattern Recognition* (pp. 130-141).

[52] Fang, J., Yi, T., Wang, X., Xie, L., Zhang, X., Liu, W., Nießner, M., & Tian, Q. (2022). Fast dynamic radiance fields with time-aware neural voxels. In *SIGGRAPH Asia 2022 Conference Papers* (pp. 1-9).

[53] Yoon, J. S., Kim, K., Gallo, O., Park, H. S., & Kautz, J. (2020). Novel view synthesis of dynamic scenes with globally coherent depths from a monocular camera. In *Proceedings of the IEEE/CVF Conference on Computer Vision and Pattern Recognition* (pp. 5336-5345).

[54] Li, Z., Chen, Z., Li, Z., & Xu, Y. (2024). Spacetime gaussian feature splatting for real-time dynamic view synthesis. In Proceedings of the IEEE/CVF Conference on Computer Vision and Pattern Recognition (pp. 8508-8520).

[55] Liu, Y. L., Gao, C., Meuleman, A., Tseng, H. Y., Saraf, A., Kim, C., Chuang, Y. Y., Kopf, J., & Huang, J. B. (2023). Robust dynamic radiance fields. In Proceedings of the IEEE/CVF Conference on Computer Vision and Pattern Recognition (pp. 13-23).